\documentclass{article}
\usepackage[utf8]{inputenc}
\usepackage[final]{neurips_2019}

\usepackage[colorlinks=true,linkcolor=blue,citecolor=black,urlcolor=blue]{hyperref}
\usepackage{natbib}
\usepackage{wrapfig}
\usepackage{graphicx}
\usepackage{subcaption}
\usepackage{amsmath}
\usepackage[scientific-notation=true]{siunitx}

\usepackage{graphicx}
\usepackage{amsmath}
\usepackage{amsthm}
\usepackage{booktabs}
\usepackage{algorithm}
\usepackage[table]{xcolor}
\usepackage{bbm}
\usepackage{comment}

\usepackage[linewidth=1pt]{mdframed}
\usepackage{tikz-dependency}
\usepackage{amsmath, graphicx, amsfonts, amssymb}
\usepackage{algorithmicx}
\usepackage{algorithm}
\usepackage{booktabs}
\usepackage[noend]{algpseudocode}
\usepackage[percent]{overpic}
\usepackage{subcaption}
\usepackage[page]{appendix}
\usepackage{color}
\usepackage{listings}
\NewEnviron{elaboration}{
\par
\begin{tikzpicture}
\node[rectangle,minimum width=\textwidth] (m) {\begin{minipage}{.98\textwidth}\BODY\end{minipage}};
\draw[dashed] (m.south west) rectangle (m.north east);
\end{tikzpicture}
}

\newcommand\Mark[1]{\textcolor{blue}{Mark: #1}}

\title{Situated Language Learning \\ via Interactive Narratives}

\date{\vspace{-5ex}}

\author{Prithviraj Ammanabrolu \\ Georgia Institute of Technology \\ \texttt{raj.ammanabrolu@gatech.edu} \And Mark O. Riedl \\ Georgia Institute of Technology \\ \texttt{riedl@cc.gatech.edu}}
\begin{document}

\maketitle

\begin{abstract}
%1500 character limit
This paper provides a roadmap that explores the question of how to imbue learning agents with the ability to understand and generate contextually relevant natural language in service of achieving a goal.
We hypothesize that two key components in creating such agents are interactivity and environment grounding, shown to be vital parts of language learning in humans, and posit that {\em interactive narratives} should be the environments of choice for such training these agents.
These games are simulations in which an agent interacts with the world through natural language---``perceiving'', ``acting upon'', and ``talking to'' the world using textual descriptions, commands, and dialogue---and as such exist at the intersection of natural language processing, storytelling, and sequential decision making. 
We discuss the unique challenges a text games' puzzle-like structure combined with natural language state-and-action spaces provides: knowledge representation, commonsense reasoning, and exploration.
Beyond the challenges described so far, progress in the realm of interactive narratives can be applied in adjacent problem domains.
These applications provide interesting challenges of their own as well as extensions to those discussed so far.
We describe three of them in detail: (1) evaluating AI system's commonsense understanding by automatically creating interactive narratives; (2) adapting abstract text-based policies to include other modalities such as vision; and (3) enabling multi-agent and human-AI collaboration in shared, situated worlds. 

\end{abstract}

\section{Introduction}
Natural language communication has long been considered a defining characteristic of human intelligence.
In humans, this communication is grounded in experience and real world context---``what'' we say or do depends on the current context around us and ``why'' we say or do something draws on commonsense knowledge gained through experience.
So how do we imbue learning agents with the ability to understand and generate contextually relevant natural language in service of achieving a goal?
%\MarkLeft{Just ask the rhetorical, ``How do we imbue...''}

Two key components in creating such agents are interactivity and environment grounding, shown to be vital parts of language learning in humans.
Humans learn various skills such as language, vision, motor skills, etc. more effectively through interactive media~\citep{Feldman2004,Barsalou2008}.
In the realm of machines, interactive environments have served as cornerstones in the quest to develop more robust algorithms for learning agents across many machine learning sub-communities.
Environments such as the Atari Learning Environment~\citep{Bellemare2013} and Minecraft~\citep{Johnson2016} have enabled the development of game agents that perform complex tasks while operating on raw video inputs, and more recently THOR~\citep{Kolve2017} and Habitat~\citep{ManolisSavva*2019} attempt to do the same with embodied agents in simulated 3D worlds.

Despite such progress in modern machine learning and natural language processing, agents that can communicate with humans (and other agents) through natural language in pursuit of their goals are still primitive. 
One possible reason for this is that many datasets and tasks used for NLP are static, not supporting interaction and language grounding ~\citep{Brooks1991,Feldman2004,Barsalou2008,Mikolov2016,Gauthier2016,Lake2017}
%\MarkLeft{Cell might not want extensive citations, but leave it in for now}.
In other words, there has been a void for such interactive environments for purely language-oriented tasks.
Building on recent work in this field, we posit that interactive narratives should be the environments of choice for such language-oriented tasks.
%\MarkRight{Might want to simplify this down to text games}
%I like having interactive narrative in here because it broadens the umbrella, lets us say things like visual games etc exist
\textbf{Interactive Narratives}, in general, is an umbrella term, that refers to any form of digital interactive experience in which users create or influence a dramatic storyline through their actions~\citep{Riedl2013}---i.e. the overall story progression in the game is not pre-determined and is directly influenced by a player's choices.
For the purposes of this work, we consider one particular type of interactive narrative, parser-based interactive fiction (or text-adventure) games---though we note that other forms of interactive narrative, including those with visual components, provide closely related challenges.

\begin{comment}
\begin{figure}[htp]
\begin{mdframed}
\begin{elaboration}
  \parbox{.99\textwidth}{
\emph{Observation:} \textbf{West of House} You are standing in an open field west of a white house, with a boarded front door. There is a small mailbox here.
}
\end{elaboration}
\begin{flushleft}
\emph{Action:} \textbf{Open mailbox}
\end{flushleft}
\begin{elaboration}
  \noindent\parbox{.99\textwidth}{
\emph{Observation:} Opening the small mailbox reveals a leaflet.
}
\end{elaboration}
\begin{flushleft}
\emph{Action:} \textbf{Read leaflet}
\end{flushleft}
\begin{elaboration}
  \noindent\parbox{.99\textwidth}{
\emph{Observation:} (Taken) "WELCOME TO ZORK!
ZORK is a game of adventure, danger, and low cunning. In it you will explore some of the most amazing territory ever seen by mortals. No computer should be without
one!"
}
\end{elaboration}
\begin{flushleft}
\emph{Action:} \textbf{Go north}
\end{flushleft}
\begin{elaboration}
  \parbox{.99\textwidth}{
\emph{Observation:} \textbf{North of House}
You are facing the north side of a white house. There is no door here, and all the windows are boarded up. To the north a narrow path winds through the trees.
}
\end{elaboration}
\end{mdframed}
\caption{An excerpt from {\em Zork1}, a typical text-based adventure game.\Mark{For this audience I wonder if an actual screenshot might be possible?}}
\label{fig:zorkexcerpt}
\end{figure}
\end{comment}
\begin{figure}
    \centering
    \includegraphics[width=.85\linewidth]{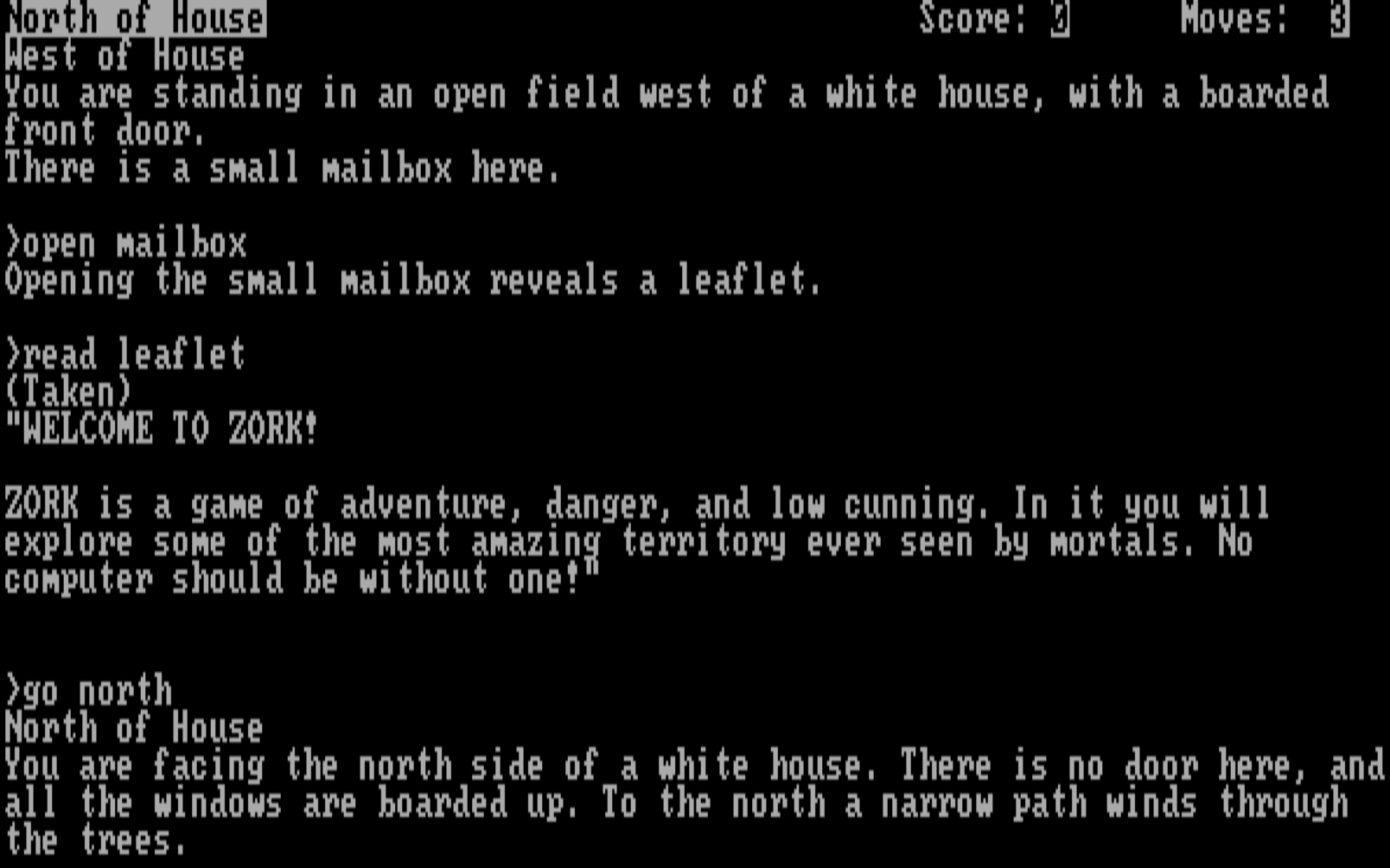}
    \caption{An excerpt from {\em Zork1}, a typical text-based adventure game.}
    \label{fig:zorkexcerpt}
\end{figure}

Figure~\ref{fig:zorkexcerpt} showcases {\em Zork}~\citep{Anderson1979}, one of the earliest and most influential text-based interactive narrative.
These games are simulations
in which an agent interacts with the world through natural language---``perceiving'', ``acting upon'', and ``talking to'' the world using textual descriptions, commands, and dialogue.
The simulations are {\em partially observable}, meaning that the agent never has access to the true underlying world state and has to reason about how to act in the world based only on potentially the incomplete textual observations of its immediate surroundings. 
They provide tractable, situated environments in which to explore highly complex interactive grounded language learning without the complications that arise when modeling physical motor control and vision---situations that voice assistants such as Siri or Alexa might find themselves in when improvising responses.
These games are usually structured as puzzles or quests with long-term dependencies in which a player must complete a sequence of actions and/or dialogues to succeed. This in turn requires navigation and interaction with hundreds of locations, characters, and objects.
The interactive narrative community is one of the oldest gaming communities and game developers in this genre are quite creative.
Put these two things together and we get very large, complex worlds that contain a multitude of puzzles and quests to solve across many different genres---everything from slice of life simulators where the player cooks a recipe in their home to Lovecraftian horror mysteries.
The complexity and diversity of topics enable us to build and test agents that go an extra step towards modeling the difficulty of situated human language communication.  

As the excerpt of the text-game in Figure \ref{fig:zorkexcerpt} shows, humans bring competencies in natural language understanding, commonsense reasoning, and deduction to bear in order to infer the context and objectives of a game. 
Beyond games, real-world applications such as voice-activated personal assistants can also benefit from advances in these capabilities at the intersection of natural language understanding, natural language generation, and sequential decision making.
These real world applications require the ability to reason with ungrounded natural language (unlike multimodal environments that provide visual grounding for language) and interactive narratives provide an excellent suite of environments to tackle these challenges.

Currently, three primary open-source platforms and baseline benchmarks have been developed so far to help measure progress in this field:
{\em Jericho}~\citep{Hausknecht2020}~\footnote{\url{https://github.com/microsoft/jericho}} a learning environment for human-made interactive narrative games; 
{\em TextWorld}~\citep{Cote2018}~\footnote{\url{https://github.com/microsoft/textworld}} a framework for procedural generation in text-games; 
and {\em LIGHT}~\citep{Urbanek2019}~\footnote{\url{https://parl.ai/projects/light}} a large-scale crowdsourced multi-user text-game for studying situated dialogue.

%\todo{scaffold}
\begin{figure}
    \centering
    \includegraphics[width=\linewidth]{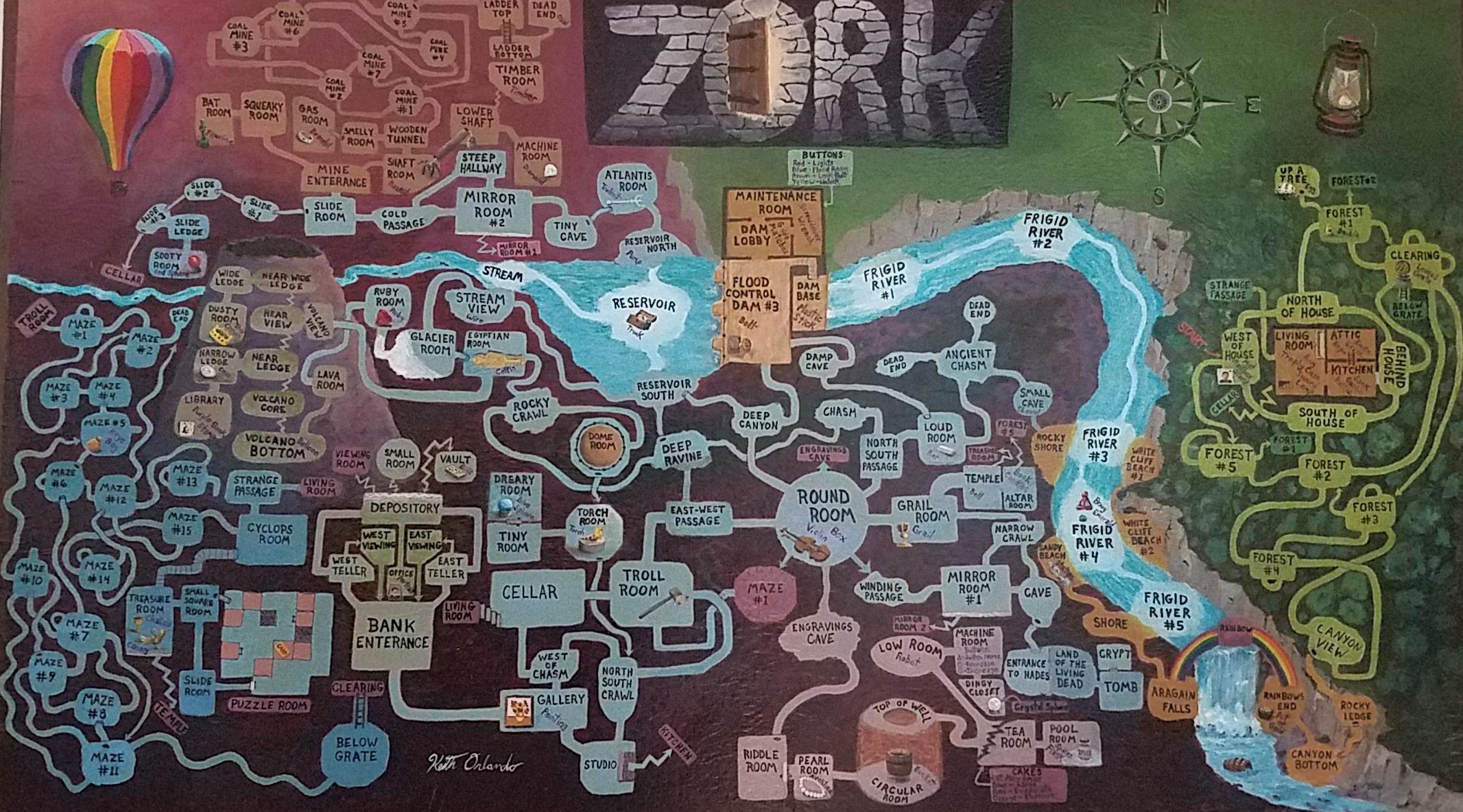}
    \caption{A map of {\em Zork1} by artist \href{https://www.reddit.com/r/zork/comments/5ximil/zork_map_mit_version_fantastic_game_i_spent_some/}{\em ion\_bond}.}
    \label{fig:zorkmapch1}
\end{figure}
\section{Challenges}
Interactive narratives exist at the intersection of natural language processing, storytelling, and sequential decision making. 
Like many NLP tasks, they require natural language understanding, but unlike most NLP tasks, Interactive narratives are sequential decision making problems in which actions change the subsequent world states of the game and choices made early in a game may have long term effects on the eventual endings.
Reinforcement Learning~\citep{Sutton1998} studies sequential decision making problems and has shown promise in vision-based~\citep{Jaderberg2016} and control-based~\citep{OpenAI2018} environments, but has less commonly been applied in the context of language-based tasks.
Text-based games thus pose a different set of challenges than traditional video games such as {\em StarCraft}.
Their puzzle-like structure coupled with a partially observable state space and sparse rewards require a greater understanding of previous context to enable more effective exploration---an implicit {\em long-term dependency} problem not often found in other domains that agents must overcome.

\subsection{Knowledge Representation}
%Structured memory

Interactive narratives span many distinct locations, each with unique descriptions, objects, and characters.
An example of a world of a interactive fiction game can be seen in Figure~\ref{fig:zorkmapch1}.
Players move between locations by issuing navigational commands like \textit{go West}. 

This, in conjunction with the inherent {\em partial observability} of interactive narratives, gives rise to the \textbf{Textual-SLAM} problem, a textual variant of Simultaneous localization and mapping (SLAM)~\citep{Thrun2005} problem of constructing a map while navigating a new environment. 
In particular, because connectivity between locations is not necessarily Euclidean, agents need to detect when a navigational action has succeeded or failed and whether the location reached was previously seen or new. 
Beyond location connectivity, it's also helpful to keep track of the objects present at each location, with the understanding that objects can be nested inside of other objects, such as food in a refrigerator or a sword in a chest.

Due to the large number of locations in many games, humans often create structured memory aids such as maps to navigate efficiently and avoid getting lost. 
The creation of such memory aids has been shown to be critical in helping automated learning agents operate in these textual worlds~\citep{Ammanabrolu2019, Murugesan2020, Adhikari2020, Ammanabrolu2020c}

\subsection{Acting and Speaking in Combinatorially-sized State-Action Spaces}
%adapt RL algos to crazy space, 
Interactive narratives require the agent to operate in the combinatorial action space of natural language.
To realize how difficult a game such as {\em Zork1} is for standard reinforcement learning agents, we need to first understand how large this space really is.
In order to solve solve a popular IF game such as {\em Zork1} it's necessary to generate actions consisting of up to five-words from a relatively modest vocabulary of 697 words recognized by Zork's parser.
Even this modestly sized vocabulary leads to $\mathcal{O}(697^5)=\num{1.64e14}$ possible actions at every step---a dauntingly-large \emph{combinatorially-sized action space} for a learning agent to explore.
In comparison, board games such as chess and Go or Atari video games have branching factors of the order of $\mathcal{O}(10^2)$.

Some text-games extend this even further by requiring agents to engage in dialogue to progress in a task, increasing the space of possibilities exponentially and bringing text environments closer to real-world situations.
An example of such an environment---designed explicitly as a research platform--- 
\begin{wrapfigure}[18]{r}{.6\textwidth}
    \vspace{-5pt}
    \centering
    \includegraphics[width=\linewidth]{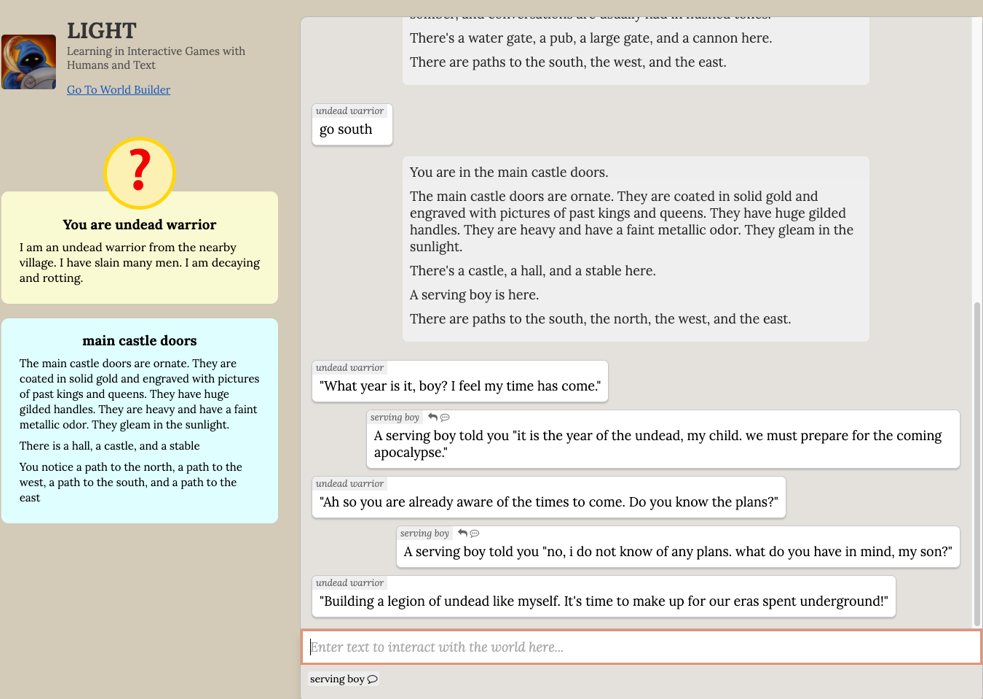}
    \caption{The {\em LIGHT}~\citep{Urbanek2019} environment.}
    \label{fig:light}
\end{wrapfigure}
is the large-scale crowdsourced fantasy text-adventure game {\em LIGHT}~\citep{Urbanek2019}, seen in Figure~\ref{fig:light}, where characters can act and talk while interacting with other characters.
It consists of of a set of locations, characters, and objects leading to rich textual worlds in addition to quests demonstrations of humans playing these quests providing natural language descriptions in varying levels of abstraction of motivations for a given character in a particular setting.

On top of the other text-game related challenges, the primary core challenge for the agent here is the recognition that dialogue can also be used to change the environment.
With dialogue, an agent can now learn to instruct or convince other characters in the world to achieve the goal for it---e.g. convince the pirate through dialogue to give you their treasure instead of just stealing it yourself.
The agent needs to learn to balance both its ability to speak as well as act in order to effectively achieve its goals~\citep{Ammanabrolu2021}.
%On top of the other text-game related challenges, the primary core challenge here is to learn how to balance speech with actions.
%\MarkRight{Might reframe this as a recognition that dialogue is a means of changing the environment because one can instructor or convince agents in LIGHT to achieve goals.}
%An interesting end result of learning to talk in these games is that an agent can now learn to convince other characters in the world to achieve the goal for it---e.g. convince the pirate through dialogue to give you their treasure instead of just stealing it yourself.

\subsection{Commonsense Reasoning}
%games are diverse, how to transfer, lifelong learning

Text-games cover a wide variety of genres, as mentioned earlier this ranges from slice of life simulators where the player makes a recipe in their home to Lovecraftian horror mysteries.
In order to effectively convey the core narrative or puzzle, 
%In essence, the agent must learn everything about the game from only its interactions with the environment.
text-adventure games make ample use of prior commonsense knowledge.
Everyday example could be something as mundane as the fact that an axe can be used to cut wood, or that swords are weapons.
Different genres also have specific knowledge attached to them that wouldn't normally be found in mundane settings, e.g. in a horror or fantasy game, we know that a coffin is likely to contain a vampire or other undead monster or that kings are royalty and must be treated respectfully.
When a human enters a particular domain, they already possess priors regarding the specific knowledge relevant to the situations likely to be encountered---this is thematic commonsense knowledge that a learning agent must acquire to ensure successful interactions.

This is closely related to the problem of {\em transfer}, the problem of acquiring and adapting these priors in novel environments through interaction.
In this sense, we can think of commonsense knowledge as priors regarding environment dynamics.
This problem space can be explored using text-based games.
What commonsense can be transferred between two different environments, for example, a horror game and a mundane slice of life game?
How do you unlearn, or choose not to apply, a piece of commonsense that no longer fits with the current world. 
What if the perceived environment dynamics change in novel ways?
E.g. some vampires actually love garlic instead of being allergic to them or you suddenly find out that bread can be made without yeast and is known as sourdough---whole new categories of recipes are now possible. 

\subsection{Exploration}
%sparse feedback, dag structure of quests in games, unwinnable states (remember the blog post)

Most text-adventure games have relatively linear plots in which players must solve a sequence of puzzles to advance the story and gain score.
To solve these puzzles, players have freedom to a explore both new areas and previously unlocked areas of the game, collect clues, and acquire tools needed to solve the next puzzle and unlock the next portion of the game.
From a Reinforcement Learning perspective, these puzzles can be viewed as bottlenecks that act as partitions between different regions of the state space.
Whereas the relatively linear progression through puzzles may seem to make the problem easier, the opposite is true. 
The bottlenecks set up a situation where agents get stuck because they do not see the right action sequence enough times to be sufficiently reinforced.
We contend that existing Reinforcement Learning agents are unaware of such latent structure and are thus poorly equipped for solving these types of problems.
%---the agents states are unable to pass through bottlenecks simply because they do not see the right action sequence enough times to be sufficiently reinforced.

Overcoming bottlenecks is not as simple as selecting the correct action from the bottleneck state. 
Most bottlenecks have long-range dependencies that must first be satisfied:
{\em Zork1} for instance features a bottleneck in which the agent must pass through the unlit {\em Cellar} where a monster known as a Grue lurks, ready to eat unsuspecting players who enter without a light source.
To pass this bottleneck the player must have previously acquired and lit the lantern.
Reaching the {\em Cellar} without acquiring the lantern results in the player reaching an {\em unwinnable state}---the player is unable to go back and acquire a lantern but also cannot progress further without a way to combat the darkness.
Other bottlenecks don't rely on inventory items and instead require the player to have satisfied an external condition such as visiting the reservoir control to drain water from a submerged room before being able to visit it.
In both cases, the actions that fulfill dependencies of the bottleneck, e.g. acquiring the lantern or draining the room, are not rewarded by the game. 
Thus agents must correctly satisfy all \emph{latent} dependencies, most of which are unrewarded, then take the right action from the correct location to overcome such bottlenecks.
Consequently, most existing agents---regardless of whether they use a reduced action space~\citep{Zahavy2018,Yuan2018,Yin2019a} or the full space~\citep{Hausknecht2020,Ammanabrolu2020c}---have failed to consistently clear these bottlenecks.
It is only recently that works have begun explicitly accounting for and surpassing such bottlenecks---using a reduced action space and Monte-Carlo Planning~\citep{Jang2021} and full action space and intrinsic motivation-based structured exploration~\citep{Ammanabrolu2020}.

\section{Applications and Future Directions}
Beyond the challenges described so far, progress in the realm of interactive narratives can be applied in adjacent problem domains.
These applications provide interesting challenges of their own as well as extensions to those discussed so far.
This section will describe three of them in detail: (1) evaluating AI system's commonsense understanding by creating interactive narratives; (2) adapting abstract text-based policies to include other modalities such as vision; and (3) enabling multi-agent and human-AI collaboration in shared, situated worlds. 

\subsection{Automated World and Quest Generation}

A key consideration in modeling communication through a general purpose interactive narrative solver is that an agent trained to solve these games is limited by the scenarios described in them.
Although the range of scenarios is vast, this brings about the question of what the agent is actually capable of understanding even if it has learned to solve all the puzzles in a particular game.
Deep (reinforcement) learning systems tend to learn to generalize from the head of any particular data distribution, the ``common'' scenarios, and memorize the tail, the rarely seen cases.
We contend that a potential way of testing an AI system's understanding of a domain is to use the knowledge it has gained in a novel way and to create more instances of that domain.

From the perspective of interactive narratives, this involves automatically creating such games---the flip side of the problem of creating agents that operate in these environments---and requires {\em anticipating} how people will interact with these environments and conforming to such expected commonsense norms to make a creative and engaging experience.
The core experience in an interactive narrative revolves the quest, consisting of the partial ordering of activities that an agent must engage in to make progress toward the end of the game.
{\em Quest generation} requires narrative intelligence and commonsense knowledge as a quest must maintain coherence throughout while progressing towards a goal~\citep{ammanabrolu2020towards}.
Each step of the quest follows logically from the preceding steps much like the steps of a cooking recipe.
A restaurant cannot serve a batch of cookies without first gathering ingredients, preparing cooking instruments, mixing ingredients, etc. in a particular sequence.
Any generated quest that doesn't follow such an ordering will appear random or nonsensical to a human, betraying the AI's lack of commonsense understanding.

Maintaining quest coherence also means following the constraints of the given game world.
The quest has to fit within the confines of the world in terms of both genre and given affordances---e.g. using magic in a fantasy world, placing kitchens next to living rooms in mundane worlds, etc.
This gives rise to the concept of {\em world generation}, the second half of the automated game generation problem.
This refers to generating the structure of the world, including the layout of rooms, textual description of rooms, objects, and characters---setting the boundaries for how an agent is allowed to interact with the world~\citep{Ammanabrolu2020a}.
Similarly to quests, a world violating thematically relevant commonsense structuring rules will appear random to humans, providing us with a metric to measure an AI system's understanding.

\subsection{Transfer across domains and modalities}
\begin{wrapfigure}[24]{r}{.5\textwidth}
    \vspace{-20pt}
    \centering
    \includegraphics[width=\linewidth]{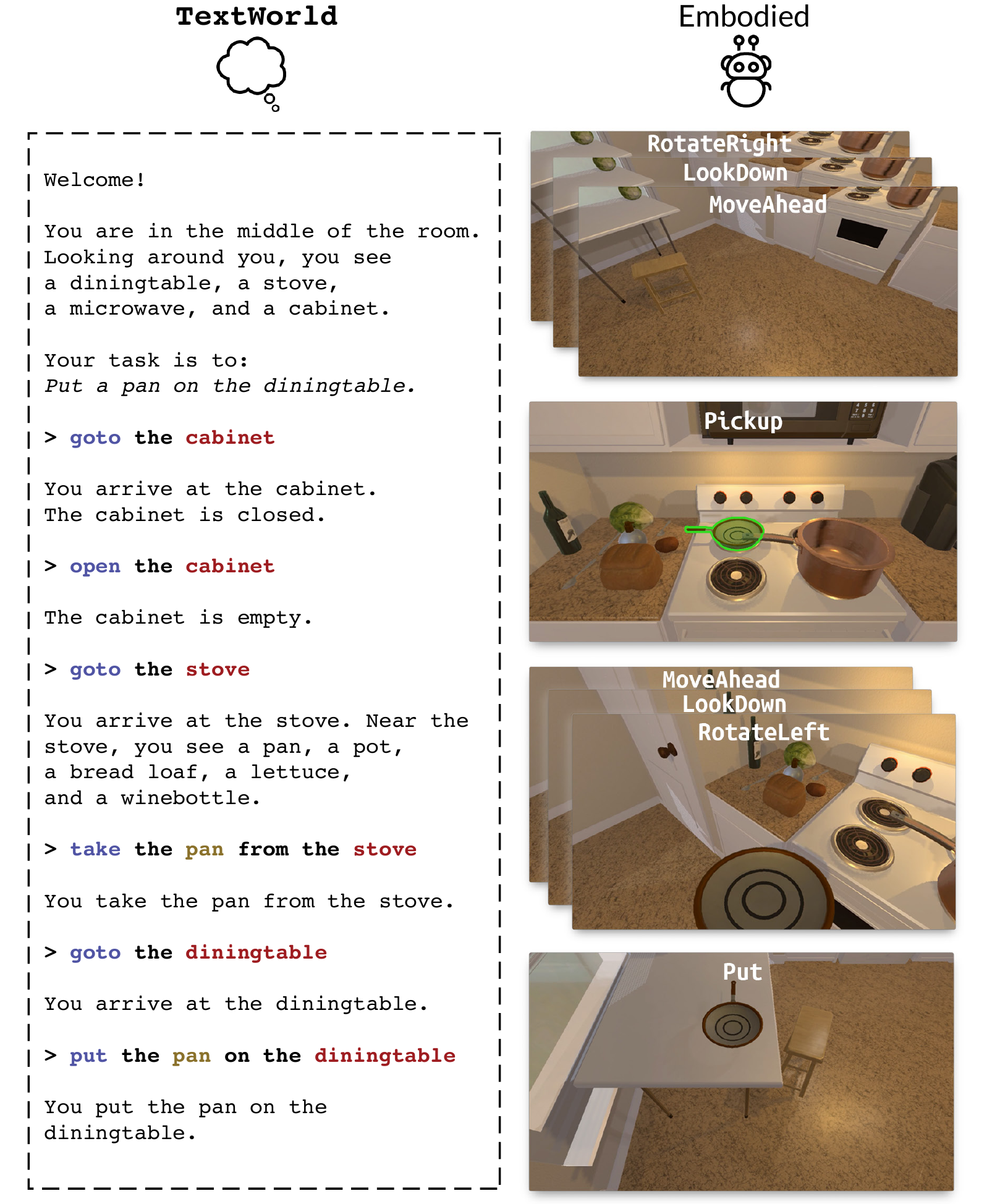}
    \caption{{\em ALFWorld}~\citep{shridhar2021alfworld}.
    %\Mark{Cell may technically require copyright permission to use this if it is from the AlfWorld paper. You may want to reach out to Shridhar et al and get permission in advance.}
    }
    \label{fig:alfworld}
\end{wrapfigure}
Many of the core challenges presented by text games manifest themselves across domains with different modalities and it may be possible to transfer progress between the domains.
Take the example of a slice-of-life walking simulator text game where the main quest is to complete a recipe as given before.
What happens when we encounter a similar situation with the added modality of vision?
Can we take the knowledge we've gained from learning a text-based policy by completing the recipe in the original text game and use that to learn how to do something similar with a visually embodied agent?
To test this idea, \citet{shridhar2021alfworld} built ALFWorld, a simulator that lets you first learn text-based policies in the ``home'' text-game TextWorld~\citep{Cote2018}, and then execute them in similarly themed scenarios from the visual environment ALFRED~\citep{ALFRED20}.
They find that commonsense priors---regarding things like common object locations, affordances, and causality---learned while playing text-games can be adapted to help create agents that generalize better in visually grounded environments.
This indicates that text games are suitable environments to train agents to reason abstractly through text which can then be refined and adapted to specific instances in an embodied setting.

\begin{wrapfigure}[16]{l}{.525\textwidth}
    \vspace{-10pt}
    \centering
    \includegraphics[width=\linewidth]{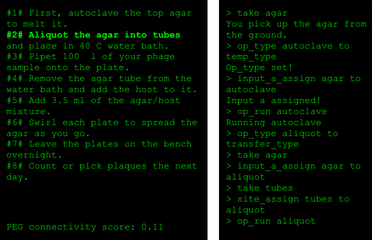}
    \caption{A wet lab protocol as a text game from the X-WLP dataset~\citep{tamari2021process}.}
    \label{fig:xwlp}
\end{wrapfigure}
Another such cross-domain transfer experiment was tested by \citet{tamari2021process},
where they collected and built X-WLP, a corpus of complex wet lab biochemistry protocols that are framed as a quest and could thus be executed via a text-game engine.
The annotations themselves are collected using a text-game-like interface, reducing overall data collection cost.
\citet{Tamari2019} discuss automatically extracting these protocols from raw lab texts and also training deep reinforcement learning agents on the resulting text-game quest.
The ability to automatically frame wet lab experiments in the form of text game quests and leverage the latest text-game agent advances to interactively train agents to perform them has implications for significantly improving procedural text understanding~\citep{levy-etal-2017-zero} and in the reproducibility of scientific experiments~\citep{Mehr101}.

\subsection{Multi-agent and Human-AI Collaboration}
Current work on teaching agents to act and speak in situated, shared worlds such as LIGHT opens the doors for exploring multi-agent communication using natural language, i.e. through dialogue.
It has been shown how to teach agents to act and talk in pursuit of a goal in this world leads to them learning multiple ways of achieve the goal: acting to do it themselves, or convincing a partner agent to do it for them.
We envision this situated learning paradigm extended to to a multi-agent setting, where there are multiple agents progressing through a world in pursuit of their own motivations that learn to communicate with each other, figuring out what others can do for them.
This gives rise to a dynamic world within the bounds of a {\em unified decision making framework}, a situation autonomous agents are likely to find themselves in.
A village led by an ambitious chief seeking expansion will expand into a town via environment dynamics, or narrative, emerging from this multi-agent communication.
Agents can further be taught which other agents they should cooperate with and which they should compete with on the basis of the alignment of their motivations.
A dragon terrorizing a kingdom and a knight may perhaps be at odds, but the kingdom's ruler will have cause to cooperate and explicitly aid the knight in slaying the dragon.
A not-so-fantastic example would be two small clothing businesses cooperating and pooling resources to compete against an encroaching large corporation.  

A human-AI collaborative system is an instance of such a multi-agent system where one or more of the agents are humans.
These works thus have direct implications for human-AI collaborative systems: from agents that act and talk in multi-user worlds, to improvisational and collaborative storytelling, and creative writing assistants for human authors.

\section{Conclusion}
{\em Interactive narratives} provide tractable, situated environments in which to explore highly complex interactive grounded language learning without the complications that arise when modeling physical motor control and vision.
The unique challenges a text games' puzzle-like structure combined with natural language state-and-action spaces provides is: knowledge representation, commonsense reasoning, and exploration.
These challenges create an implicit {\em long-term dependency} problem not often found in other domains that agents must overcome.
Text-based games thus pose a different set of challenges than traditional video games such as {\em StarCraft}.
Beyond the challenges described so far, we have seen how progress in the realm of interactive narratives can be applied in adjacent problem domains, specifically: (1) structured environment creation; (2) transfer to other modalities and domains; and (3) enabling multi-agent and human-AI collaboration in shared, situated worlds. 

\section*{Acknowledgements}
We thank Matthew Hausknecht, Xingdi Yuan, and Marc-Alexandre Côté of Microsoft Research for useful discussions on text games and their work on the Jericho and TextWorld platforms.
Likewise, thanks to Jack Urbanek, Margaret Li, Arthur Szlam, Tim Rockt{\"a}schel, and Jason Weston of Facebook AI Research for their efforts and guidance in the work on the LIGHT framework.
We also would like to thank the corresponding authors Mohit Shridar of the University of Washington and Ronen Tamari of the Hebrew University of Jerusalem for discussions regarding their respective works ALFWorld and X-WLP and the images within reproduced accordingly.
\bibliography{bib}

\end{document}